\documentclass[letterpaper, 10 pt, conference]{ieeeconf}  

\IEEEoverridecommandlockouts                              

\makeatletter
\let\NAT@parse\undefined
\makeatother

\usepackage{graphics}

\usepackage{caption}
\usepackage{graphicx}
\usepackage{grffile} 

\usepackage{amsmath} 
\usepackage{amssymb}
\usepackage{color}
\usepackage{bm}
\usepackage{url}

\usepackage[noend]{algpseudocode}
\usepackage{algorithm}
\usepackage{makecell}
\usepackage{balance}
\usepackage{pifont} 
\usepackage[square, comma, numbers,sort]{natbib}
\usepackage{multirow}

\let\labelindent\relax
\usepackage{enumitem}
\usepackage[table,dvipsnames]{xcolor}
\usepackage[pagebackref=true,breaklinks=true,colorlinks,bookmarks=false]{hyperref}

\hypersetup{
linkcolor=BrickRed
,citecolor=Green
,filecolor=Mulberry
,urlcolor=NavyBlue
,menucolor=BrickRed
,runcolor=Mulberry
,linkbordercolor=BrickRed
,citebordercolor=Green
,filebordercolor=Mulberry
,urlbordercolor=NavyBlue
,menubordercolor=BrickRed
,runbordercolor=Mulberry
}

\newcommand{\acronym}{EgoPAT3Dv2}

\def\etal{et al. }

\title{\LARGE \bf
EgoPAT3Dv2: Predicting 3D Action Target from 2D Egocentric Vision for Human-Robot Interaction\vspace{-0.3cm}
}
\author{Irving Fang$^{1,*}$, Yuzhong Chen$^{1,*}$, Yifan Wang$^{1,*}$, Jianghan Zhang$^{1, \dagger}$, Qiushi Zhang$^{1, \dagger}$, Jiali Xu$^{1, \dagger}$, Xibo He$^{1}$,\\Weibo Gao$^{2}$, Hao Su$^{2}$, Yiming Li$^{1}$, Chen Feng\textsuperscript{1,\ding{41}}\\
{\tt\small \url{https://ai4ce.github.io/EgoPAT3Dv2/}}
\thanks{$^{1}$New York University, Brooklyn, NY 11201, USA}
\thanks{$^{2}$North Carolina State University, Raleigh, NC 27695, USA}
\thanks{$^{*, \dagger}$Equal contributions.}
\thanks{\ding{41} Corresponding author ({\tt\small \href{mailto:cfeng}{cfeng@nyu.edu}}). This work is supported by NSF Grant 2026479, and by NYU IT High-Performance Computing resources, services, and staff expertise.}
}

\IEEEaftertitletext{
\begin{center}
    \vspace{-10mm}
    \centering
    \captionsetup{type=figure}
    \includegraphics[width=\textwidth]{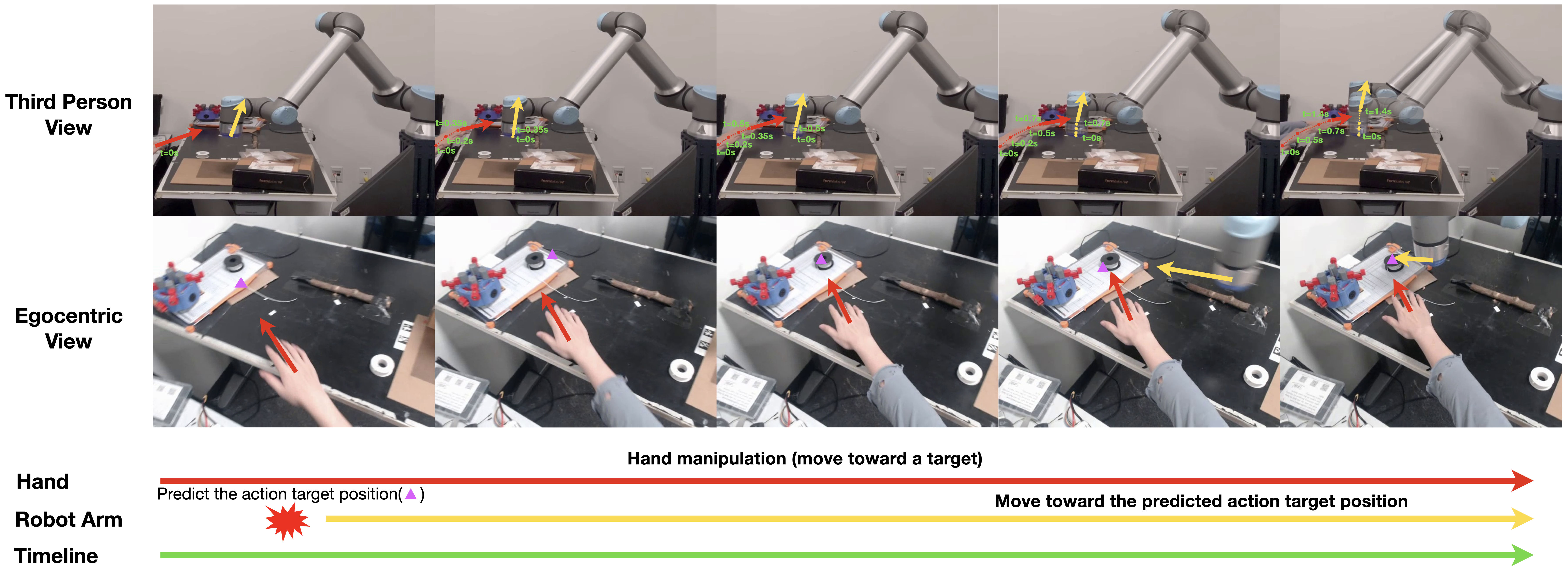}
    \captionof{figure}{\textbf{Real-World Demonstration of \acronym}. A human wearing a helmet camera manipulates objects in a shared workspace with a UR10E cobot. The cobot tries to reach the anticipated 3D action target with the shortest Cartesian path. }
    \label{fig:fig1}
    \vspace{-0.2cm}
\end{center}%
}

\begin{document}

\maketitle
\thispagestyle{empty}
\pagestyle{empty}

%%%%%%%%%%%%%%%%%%%%%%%%%%%%%%%%%%%%%%%%%%%%%%%%%%%%%%%%%%%%%%%%%%%%%%%%%%%%%%%%

\begin{abstract}
% Being able to make online predictions of the hand's 3D target coordinate in a manipulation task in egocentric videos can be helpful for many human-robot interaction scenarios. While previous works often focus on semantic classification of action or 2D target region prediction, we believe that precise 3D anticipation can enable more versatile downstream robotics tasks, especially given the increasingly widespread commercialization of headset devices in recent years. In this work, we double the size of EgoPAT3D, the only egocentric 3D target location dataset, and increase its diversity for better generalization. We significantly improve the effectiveness and usability of the baseline algorithm for 3D target coordinate prediction. Instead of requiring 3D point clouds and IMU input, our new algorithm can use just RGB images to achieve even better prediction results. We also deploy our new baseline algorithm on a real-world robot to demonstrate a simple human-robot interaction task.

% In human-robot interaction (HRI), the ability to precisely predict a hand's three-dimensional (3D) target coordinates during a manipulation task in egocentric videos holds great promise. 
A robot's ability to anticipate the 3D action target location of a hand's movement from egocentric videos can greatly improve safety and efficiency in human-robot interaction (HRI).
While previous research predominantly focused on semantic action classification or 2D target region prediction, we argue that predicting the action target's 3D coordinate could pave the way for more versatile downstream robotics tasks, especially given the increasing prevalence of headset devices. This study expands EgoPAT3D, the sole dataset dedicated to egocentric 3D action target prediction. We augment both its size and diversity, enhancing its potential for generalization. Moreover, we substantially enhance the baseline algorithm by introducing a large pre-trained model and human prior knowledge. Remarkably, our novel algorithm can now achieve superior prediction outcomes using solely RGB images, eliminating the previous need for 3D point clouds and IMU input. Furthermore, we deploy our enhanced baseline algorithm on a real-world robotic platform to illustrate its practical utility in straightforward HRI tasks. The demonstrations showcase the real-world applicability of our advancements and may inspire more HRI use cases involving egocentric vision. All code and data are open-sourced and can be found on the project website.
\vspace{-0.3cm}
\end{abstract}

\section{Introduction}
To make robots more viable in our daily lives, intelligent and safe human-robot interaction (HRI) is essential. In the past, much work has been done to make robots' motion more legible and expressive to humans \cite{Dragan2013, Avdic2021} and make robots provide more intuitive visual feedback \cite{Cha2016}. 
Equally important in HRI is the ability of robots to anticipate human actions and adapt their own accordingly.
While many robotics research on human action anticipation adopt either a third-person view camera \cite{mainprice2013} or a camera mounted on the robot \cite{Schydlo2018}, 
addressing it using egocentric vision, namely visual input from the human's perspective, enjoys great potential and unique benefits for HRI due to the increasing prevalence of low-cost egocentric cameras (e.g., in mixed reality headsets or lifelogging devices) and the rich information they capture on both the environment and the human egomotion~\cite{Marina-Miranda2022}.

As a common task setup in HRI, object manipulation in a workspace shared by humans and robots is our focus in this work. Previous egocentric action anticipation research often studies 2D target region prediction \cite{100DOH, EgoTent}, trajectory forecasting \cite{BaoUSST_ICCV23}, or the prediction on video of fixed length \cite{leejang_iros2017}. To fill the gap between those works and real-world manipulation HRI, we need \textit{online predictions of 3D target coordinates} on variable-length videos. This results in our previous work on \textit{EgoPAT3D}~\cite{li2022egocentric} which provides the first dataset and baseline method capable of such 3D forecasting. 
% Because of our focus on manipulation, this excludes works focusing on walk trajectory forecasting, such as \cite{park16}. 

However, \textit{EgoPAT3D} has its limitations. First, there is a lack of a real-world HRI demonstration (unlike ours in Figure~\ref{fig:fig1}), which is critical to justify our research efforts into this 3D target coordinate prediction problem, and to inspire the future transfer of such methods to wearable robots and robotic prostheses. Second, the requirement of 3D inputs (point clouds or depth images) leads to a \textit{bulky wearable} (a helmet mounted with a Kinect Azure sensor) that is disadvantageous in practice. Furthermore, we find \textit{image-only methods more desirable} than using point cloud and IMU readings together. Point clouds are generally less accessible than simple RGB images. IMU data also increases sensing costs. Finally, we believe the diversity of the original dataset can be increased to boost its potential for better \textit{real-world generalization}. Addressing those limitations leads to the following contributions of \textbf{\acronym{}} in this work:

\begin{enumerate}
    \item We propose a better algorithm exploiting our priors about human hand movement to achieve significant 3D prediction accuracy improvement while using only RGB image input without 3D point clouds and IMU readings.
    \item We double the size of the original dataset by introducing more diverse background scenes and people of different skin complexions, hoping to make algorithms trained on this dataset more generalizable in the real world.
    \item We deploy our algorithm to a real cobot and enable it to perform some human-robot interaction tasks, such as reaching the predicted action target with the shortest Cartesian path and proactively avoiding human action in a shared workspace.
\end{enumerate}

\begin{figure*}
    \centering
    \includegraphics[width=\textwidth]{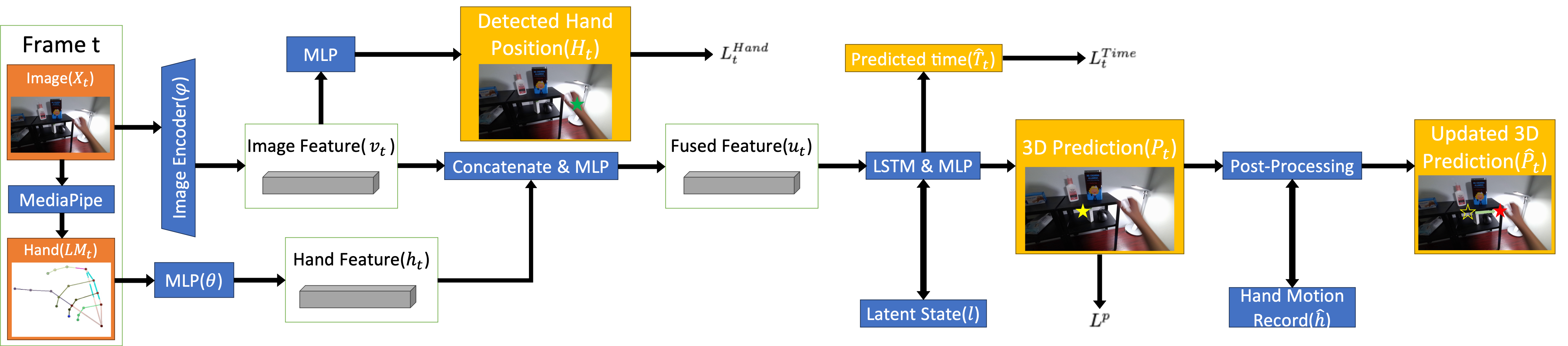}
    \caption{\textbf{Algorithm Workflow.} Visual and hand features are extracted from RGB images and fused with an MLP. The fused feature is fed into an LSTM to produce the initial prediction, which is then adjusted by post-processing that considers our prior knowledge about manipulation. Note that LSTM and Post-Processing both rely on previous frames' information.}
    \label{fig:workflow}
    \vspace{-0.5cm}
\end{figure*}
\vspace{-0.15cm}
\section{Related Work}
\subsection{\textbf{Human-Robot Interaction with Egocentric Human Action Anticipation.}}
\vspace{-0.15cm}
Traditionally, human-robot interaction (HRI) research focuses on optimizing the robot's physical movement \cite{Dragan2013, Szafir2014, mainprice2010, Zhou2017} so that the robot's actions and execution timings are more legible and expressive to humans around them. Alternatively, researchers design various visual feedback from robots to facilitate safer and more intuitive human-robot interaction \cite{Cha2016, Song2018, Szafir2014}. However, due to algorithmic and computing constraints, especially when it comes to processing high-dimensional visual data, the ability of the robot to classify or anticipate human actions before its motion generation is a relatively under-served area of research.

With the recent reviving interest in machine learning, we started to see more works focusing on integrating human action anticipation with HRI research. \cite{mainprice2013, mainprice2016} show that early prediction of human action using a Gaussian mixture model or inverse optimal control helps generate safer motion in a shared workspace. In \cite{Hawkins2013}, a Bayesian network is designed to consider human wait time in a collaborative bin manipulation task. \cite{Schydlo2018} uses an encoder-decoder recurrent neural network to predict action sequences to maximize reward in a human-robot collaborative assembly setting. While these works show promising results, the perception is often done on an overhead camera or a camera mounted on the robot.

Recently, more HRI research started to look into human egocentric vision. In \cite{Kim2019}, Kim \etal design a soft wearable robot that can understand human actions through visual input from smart glasses. Planamente \etal in \cite{Planamente2023} propose a multi-modal neural network to predict diverse human actions under changing environments. \cite{Kutbi2020} investigates the usability of a wheelchair controlled by human intention anticipation in egocentric vision. Marina-Miranda and Traver \cite{Marina-Miranda2022} use egocentric vision as a surrogate for head and eye gestures and accurately anticipate the wearer's action. \cite{Song2016, Ji2017} classify human hand gestures in egocentric vision to give corresponding commands to robot systems. \cite{leejang_iros2017} is very similar to our workflow. However, because their work aims to automatically turn unlabelled human collaboration videos into training data for imitation learning on simple human-robot collaboration, the network only predicts a 2D region where the hand will be after a fixed time interval (i.e., 1 second). Meanwhile, our network incorporates a recurrent network to deal with the uncertain length of the human action sequence to produce 3D target coordinates. So far, most of the existing literature in HRI produces semantic-level classification or prediction with egocentric vision, while our work predicts a 3D coordinate of human action target, potentially allowing more precise human-robot interactions.

While egocentric vision can be closely woven with Augmented Reality (AR) \cite{Liang2015ARIH}, our work does not involve AR technologies. For a more thorough review of AR in human-robot interaction, we refer our readers to this survey \cite{Suzuki2022}. 

\vspace{-0.15cm}
\subsection{\textbf{Egocentric Vision and Datasets.}}
\vspace{-0.15cm}
Egocentric vision is not only adopted for human action anticipation in the context of HRI. It has also proven to be useful in navigation \cite{Ohn-Bar2018}, trajectory planning \cite{park16, Bertasius2018}, hand pose estimation and segmentation \cite{ohkawa:access21, ohkawa:cvpr23} and scene understanding \cite{nagarajan2022egoenv, tan2024egodistill}. Due to the unlimited potential, we refer our readers to the following survey papers for a more thorough investigation of egocentric vision's application on video summarization \cite{delMolino2017}, hand analysis \cite{Bandini2020}, future prediction \cite{RODIN2021} and futuristic use cases \cite{plizzari2023outlook}.

While there are several existing works on human action anticipation in egocentric vision, they either produce semantic level prediction with action labels \cite{liu2020}, hand or walk trajectory \cite{liu2022joint, BaoUSST_ICCV23} or region \cite{leejang_iros2017, liu2022joint}, or focusing on predicting a future after a fixed interval \cite{liu2022joint, leejang_iros2017}. Our method focuses on predicting a 3D action target coordinate given an online streaming video with variable length.

There also exists a large amount of egocentric datasets. EPIC-KITCHENS \cite{Damen2021PAMI} and EGTEA Gaze+ \cite{li2023} facilitate research \cite{Furnari2021} on action anticipation, usually on a semantic level. Datasets like EPIC-TENT \cite{EgoTent} or 100DOH\cite{100DOH}, on the other hand, provide annotation for region prediction, which is a 2D bounding box that will indicate the hand's intended location in the future. Ego4D \cite{ego4d} allows action anticipation, region prediction, and walk trajectory prediction at the same time. Unlike previous work, which usually focuses on action labels or 2D prediction, EgoPAT3D \cite{li2022egocentric}, along with our enhancement on it, provides 3D target coordinates on videos of varying lengths. For a more thorough review of the different modalities and annotations that different Egocentric datasets provide, we refer our readers to \cite{li2022egocentric, plizzari2023outlook}.
\vspace{-0.2cm}

\section{Method}\label{sec:method}
\vspace{-0.2cm}
\subsection{\textbf{Overview}}
This section will describe the mathematical notation we will use to rigorously define the egocentric 3D action target prediction problem and the specific details of our improved algorithm \textit{\acronym{}}. Please refer to Sec. \ref{subsubsec:exp1_discuss} for more discussions and comparisons on some design choices.
\subsection{\textbf{Problem Formulation and Notation}}
\vspace{-0.15cm}
We briefly review the problem formulation from the original EgoPAT3D~\cite{li2022egocentric}. A video clip $\mathbf{C}$ of length $T$ is said to be made of $T$ consecutive frames $\{X_1, X_2,\cdots, X_T\}$. We want to predict the 3D target location $P_t^{gt}\in \mathbb{R}^3$ at each frame $t$. 
% Namely, we want to predict the hand's $(x, y, z)$ when the manual manipulation task is finished. 
Namely, we want to predict the hand's target location $(x, y, z)$ in the manual manipulation task.

From the machine learning perspective, we define our algorithm as a function $f$ that takes in all the frames up until the current frame $X_t$ and produces a regression output $\hat{P_t}$. We aim to find a $f$ so that $\hat{P_t} = f(X_{1:t})$ will be as close to $P_t^{gt}$ as possible, which is in the coordinate system of $X_t$.

Note that the target location is acquired at the very last frame of the video clip, but its value is transformed to the coordinate system of each video frame by using the transformation matrices 
computed from incremental ICP as explained in the original EgoPAT3D~\cite{li2022egocentric}.
% acquired with an IMU in the data collection equipment.

It is important to note that, in order to make the algorithm more useful in real-world applications, we restrict our algorithm to operating in online mode. When predicting $y$ at frame $X_t$, the algorithm only has access to the frames from the start up until the current frame $\{X_1, X_2, \cdots, X_t\}$, but not all the frames in the future. Namely, the algorithm does not have access to future frames $\{X_{t+1}, X_{t+2}, \cdots, X_{T}\}$. 

Finally, in this formulation, the frames $\{X_1, X_2, \cdots, X_T\}$ can be in various modalities. It can be point cloud, IMU readings, RGB pictures, etc or a combinations of the aforementioned modalities. We will be specific about the modality when we compare different algorithms.
\vspace{-0.15cm}
\subsection{\textbf{Improved Baseline Algorithm}}
\vspace{-0.15cm}
We improve the baseline method in EgoPAT3D~\cite{li2022egocentric} so it can perform online 3D target coordinate prediction with only RGB inputs. At each frame, the algorithm uses two backbone networks to extract \textbf{visual} and \textbf{hand} features. A feature fusion MLP network then fuses the features and passes them into a Long-Short Term Memory (LSTM) network \cite{LSTM} to generate 3D coordinate predictions for a continuous time sequence. The generated 3D target prediction for every frame is then post-processed before we acquire our final predictions. The workflow can be seen in Fig. \ref{fig:workflow}.

\subsubsection{\textbf{RGB and Hand Feature Encoding}}
We employ ConvNeXt\_Tiny \cite{ConvNext} (denoted by $(\psi)$) pre-trained on ImageNet-1K to extract visual features $v_t = \psi (X_t)$ from each RGB frame. The weights of it are not frozen. The choice of ConvNeXt\_Tiny is discussed in more detail in Sec. \ref{dis:backbone}. Hand landmarks ${LM^1_t, LM^2_t \cdots LM^{21}_t}$ are firstly extracted by the Hand API from Google's MediaPipe \cite{MediaPipe}. The underlying model is Google's proprietary technology. If no hand is detected, then all landmarks are set to 0. A multi-layer perceptron (MLP) denoted by $(\phi)$ is then used to encode hand landmarks to features $h_t = \phi (LM^{21stack}_t)$. After the feature encodings, the two features were concatenated and fed into another MLP to obtain the fused feature $u_t = MLP(cat(v_t,h_t))$ for a single frame.

\subsubsection{\textbf{Online 3D Target Prediction}}
We use a 2-layer LSTM to process the fused feature. The steps to handle LSTM outputs are similar to the original EgoPAT3D baseline \cite{li2022egocentric}. We divide a 3D space into grids of dimension $1024\times1024\times1024$ and aim to generate a confidence score for each grid. The choice of granularity at 1024 is empirically supported in \cite{li2022egocentric}. We used three separate MLPs to process the output of the LSTM and obtain the confidence scores in three dimensions. For example, without loss of generality, for dimension $x$ at frame $t$, let $g \in \mathcal{R}^{1024}$ denote all the grids in the $x$-dimension, where we normalize the coordinates of each grid to be in [-1, 1]. The score vector $s^x_{t} \in \mathcal{R}^{1024} $ is computed by $s^x_{t} = MLP_X(LSTM(u_t,l_{t-1}))$, where $l_{t-1}$ is the learned hidden representation and $l_0$ is set to be 0. A binary mask $m_{t}^x \in \mathcal{R^1024}$ is used to remove the value for all the grids where the confidence is less than a threshold $\gamma$. Let $s^x_{t}[i]$, $m^x_{t}[i]$ denote the score and mask for the $i$-th grid, we have that:
\begin{equation}
m^x_{t}[i] =
\begin{cases}
 1, & i \in {j | s^x_{t}[j] > \gamma} \\
 0, & i \in {j | s^x_{t}[j] \leq \gamma} \\
\end{cases} 
\label{eq:prediction}
\notag
\end{equation}
The masked score is then calculated by $\hat{s}^x_{t} = m^x_{t} \odot s^x_{t}$ where $\odot$ denotes the element-wise dot product. Then, we can get the estimated target position value for dimension $x$ at frame $t$ as: 
\begin{equation}
    x_t \in \mathcal{R} = (\hat{s}^x_{t})^Tg \notag
\end{equation}

\subsubsection{\textbf{Post-Processing}}\label{subsubsec:post_processing}
We conduct post-processing for each result produced by the LSTM to incorporate human prior knowledge. The specific reasoning behind this design choice is further explained in Sec. \ref{dis:post-processing}. For each frame $t$, we choose the coordinate of the landmark that marks the end of the index finger to be the 2D hand position $\Acute{h}_t$. The predicted 3D target position $P_t$ (in meter) was transformed into 2D position $\Acute{P_t}$ in pixel values with the help of camera intrinsic parameter $K$ and image resolution (4K in our case). We ignore the depth information in this transformation. We calculate the hand position offset between each frame by $\hat{h}_t = ||\Acute{h}_t - \Acute{h}_{t-1}||_2$ and keep track of the max historical hand position offset $\Bar{h}_t = max(\hat{h}_t) \text{ for } i < t$. The final 2D position is calculated as $\Bar{P_t} = \Acute{P_t}*\dfrac{\hat{h}_t}{\Bar{h}_t} + \Acute{h}_t*(1-\dfrac{\hat{h}_t}{\Bar{h}_t})$. The 2D result is then transformed back to a 3D position $\hat{P_t}$ with the pre-transformed depth, again with camera intrinsic parameter $K$ and image resolution, to serve as the final prediction.

\vspace{-0.15cm}
\subsection{\textbf{Improved Loss Function}}\label{subsec:loss}
Our loss is a modification from the truncated weighted regression loss (TWRLoss) proposed in EgoPAT3D \cite{li2022egocentric}. The TWRLoss $L^p$ directly calculates the loss between ground truths and predictions. Additionally, We incorporate two new losses \textbf{Hand Position Loss} $L^{Hand}$ and \textbf{Time Loss} $L^{Time}$. They do not directly supervise the difference between predictions and ground truths but instead aim to incorporate human prior knowledge about manipulation. 
The overall loss of our training paradigm can be written as $L = \sum_{t=1}^T w_t(L^p_{t} + \delta (L_t^{Hand} + L_t^{Time}))$ where $w_t$ is a linear weight from 2 to 1 with respect to time $t$, and $\delta$ is a hyperparameter that acts as a weight for the two new losses. More details about the design choices are discussed in Sec. \ref{dis:losses}.

\subsubsection{\textbf{Hand Position Loss}}
We want the visual feature extractor to focus more on the hand, so we introduce a task for the feature extractor and an additional MLP $H_t = MLP(v_t)$ to predict the hand position for each frame. Only frames with a hand detected will be included in the hand position loss: \begin{equation}
L_t^{Hand} =
\begin{cases}
 (H_t - \Acute{h}_t)^2, & \text{hand in frame $t$} \\
 0, & \text{hand not in frame $t$} \\
\end{cases} 
\label{eq:hand}
\notag
\end{equation}

\subsubsection{\textbf{Time Loss}}
We want the LSTM to be able to differentiate between early stages and late stages without introducing hardcoded positional encoding that relies on knowing the length of the clip beforehand, so we introduce another task for LSTM and an additional MLP $\hat{T}_t = MLP(LSTM(u_t,l_{t-1}))$ to predict where the current frame is relative to the whole clip $\Bar{T}_t = \dfrac{t}{T}$. The time loss $L_t^{Time}$ is calculated by $L_t^{Time} = (\hat{T} - \Bar{T}_t)^2$.

\section{Experiment}

\begin{table*}[t]
\scriptsize
  \centering
      \caption{\textbf{Prediction error (cm) of different models on the EgoPAT3D seen scenes.} The lower, the better. Post denotes post-processing. Hand denotes hand position input into the model. $L^{Hand}$ denotes the Hand Position Loss and $L^{Time}$ denotes the Time Loss. \textcolor{BrickRed}{Red}, \textcolor{green!60!black}{green}, and \textcolor{blue}{blue} fonts denote the top three performance.}
      \vspace{-0.1cm}
    \resizebox{\textwidth}{!}{\begin{tabular}{c|c|c|c|ccccc|ccccc}
	\hline
	\multirow{2}{*}{\textbf{Description}}& \multirow{2}{*}             {\textbf{Method}}&\multirow{2}{*}                   {\textbf{Modality}}&\multirow{2}{*}{\textbf{Overall}} &\multicolumn{5}{c|}{\textbf{Early prediction}} &\multicolumn{5}{c}{\textbf{Late prediction}} \\
	& & & &10\% & 20\%  & 30\% & 40\%& 50\% &60\% & 70\%  & 80\% & 90\%  & 100\% \\
        \hline
        Non-Learning & Random & N/A & 49.44 & 50.83 & 47.43 & 47.74 & 49.01 & 47.99 & 50.06 & 47.30 & 48.67 & 50.01  & 48.45 \\
        
        \hline

        \multirow{2}{*}{Baseline} & EgoPAT3D\_VF & Point Cloud & 18.75 & 23.45 & 21.73  & 20.11 & 18.71 & 17.52 & 16.65 & 16.15 & 16.02 & 15.97  & 16.15 \\

        & EgoPAT3D & Point Cloud, IMU & 16.70 & 20.76 & \textcolor{blue}{19.15}  & \textcolor{BrickRed}{17.85} & \textcolor{BrickRed}{16.84} & \textcolor{blue}{15.72} & 15.01 & 14.54 & 14.29 & 14.23  & 14.35 \\

        \hline

         \multirow{10}{*}{Variants}& ConvNeXt\_Only & \multirow{9}{*}{RGB} & 15.90 & 20.63 & 19.65  & 18.84 & 17.69 & 16.02 & 14.34 & 12.67  & 11.13 & 10.11  & 9.79 \\
        
        & ResNet50\_Only &  & 16.39 & 20.69 & 19.82 & 18.92 & 17.73 & 16.09 & 14.71 & 13.33 & 12.30 & 11.78 & 11.67 \\
        
         & Post+Hand &  & 15.59 & 20.81 & 19.78 & 19.02 & 17.96 & 16.28 & 14.34 & 12.23 & 10.43 & 8.20 & 7.06\\
         
         & Post+$L^{Hand}$+$L^{Time}$ &  & 15.66 & 21.72 & 20.58 & 19.68 & 18.33 & 16.22 & \textcolor{green!60!black}{13.79} & \textcolor{BrickRed}{11.46} & \textcolor{BrickRed}{9.59} & \textcolor{BrickRed}{7.50} & \textcolor{BrickRed}{6.59} \\
         
         & Hand+$L^{Hand}$+$L^{Time}$ &  & \textcolor{green!60!black}{15.41} & \textcolor{green!60!black}{20.38} & \textcolor{green!60!black}{19.09} & \textcolor{blue}{18.36} & \textcolor{green!60!black}{17.38} & \textcolor{green!60!black}{15.71} & \textcolor{blue}{13.96} & 12.11 & 10.42 & 9.32 & 8.82\\

         & Post+Hand+$L^{Hand}$& & 15.63 & \textcolor{blue}{20.56} & 19.81  & 19.13 & 18.27 & 16.49 & 14.43 & 12.24  & 11.38 & 8.10  & \textcolor{blue}{6.94} \\
        
        & Post+Hand+$L^{Time}$& & 15.60 & 20.90 & 19.65 & 18.86 & 17.89 & 16.35 & 14.39 & 12.30 & 10.51 & 8.22 & 7.08 \\
         & ResNet101\_Replace &  & \textcolor{blue}{15.54} & 20.98 & 19.95 & 19.33 & 18.22 & 16.28 & \textcolor{blue}{13.96} & \textcolor{blue}{11.70} & \textcolor{blue}{9.90} & \textcolor{blue}{7.67} & 7.05 \\

         & EgoPAT3Dv2\_Transformer & & 16.09 & 20.73 & 19.81 & 19.12 & 18.37 & 16.91 & 15.12 & 13.11 & 11.49 & 9.16 & 8.14 \\
         \cline{2-14}
        & EgoPAT3Dv2\_Full & Point Cloud, IMU, RGB & 16.19 & 21.22  & 19.50 & 18.43 & 17.53 & 16.25 & 14.81 & 13.34 & 12.26 & 10.80  & 10.09 \\
         \hline
         \acronym{} & \acronym & RGB & \textcolor{BrickRed}{14.97} & \textcolor{BrickRed}{20.36} & \textcolor{BrickRed}{19.07} & \textcolor{green!60!black}{18.34} & \textcolor{blue}{17.43} & \textcolor{BrickRed}{15.66} & \textcolor{BrickRed}{13.68} & \textcolor{green!60!black}{11.57} & \textcolor{green!60!black}{9.67} & \textcolor{BrickRed}{7.50} & \textcolor{green!60!black}{6.60} \\

        \hline
  \end{tabular}}
  \label{tab:better_algo_tab}
  \vspace{-0.4cm}
\end{table*}
We conducted two separate experiments. 
\begin{itemize}
    \item \textbf{Experiment 1:} Algorithms are trained and tested on the EgoPAT3D dataset with the same protocol as in the EgoPAT3D paper \cite{li2022egocentric}.
    \item  \textbf{Experiment 2:} Algorithms are trained and tested on our enhanced dataset which is to be explained in Sec.~\ref{sec:dataset}.
\end{itemize}

\vspace{-0.2cm}
\subsection{\textbf{Experiment Setup}}
\subsubsection{\textbf{Dataset}} \label{subsubsec:dataset}

\textbf{For Experiment 1}, we follow the exact same dataset preparation process in the EgoPAT3D paper\cite{li2022egocentric}. The models are trained and validated with data from five scenes. The test set is divided into \textbf{seen} and \textbf{unseen}. The seen part of the test set is made of unused data from the five training scenes. The unseen part is from 6 unseen scenes. While we cap the training clips at 25 frames, the validation and test sets have no frame number limit. Note that we made a few minor corrections to the ground truth label in the original dataset. After the corrections, the baseline from \cite{li2022egocentric} performs better than in the original paper.

\textbf{For Experiment 2}, we added nine new scenes to the training, validation, and seen test set and two new scenes to the unseen test set. Again, the clips from the validation and test sets have no length limit, while the clips in the training set are capped at 25 frames.

\subsubsection{\textbf{Implementation Details}} \label{subsubsec:implementation}
The baselines are trained and tested with the same hyperparameters open-sourced by the authors of EgoPAT3D \cite{li2022egocentric}. However, we modified the distributed training used in the scripts from PyTorch's DataParallel to Distributed DataParallel. The baseline model performs better than the original one because of different behaviors \cite{li_zhu_ddp_tutorial} in Distributed DataParallel, such as gradient gathering. All our \textit{\acronym{}} models are trained with the Adam optimizer \cite{Adam}. The learning rate is $10^{-4}$, and the weight decay rate is $10^{-5}$, with no hyperparameter search conducted. We deployed our training to four RTX 8000 GPUs, with a batch size of 8 on each, 32 in total. MediaPipe's hand detection and tracking confidences are all set to 0.5. The weight $\delta$ for the two new losses is set to 0.1.

All models are trained with the same set of three different random seeds and with PyTorch's \texttt{deterministic=True} and \texttt{benchmarking=False} to ensure maximal reproducibility. The results are the average of the three trials.

\subsection{\textbf{Quantitative Results and Discussions}}
Every test clip is evenly divided by frames into ten stages during evaluation. For clips that cannot be evenly divided by ten, more frames are allocated toward the early stages. As a result, errors in the early stages will have a slightly bigger impact on the average error.
\subsubsection{\textbf{Experiment 1}} \label{subsubsec:exp1_discuss}
Table \ref{tab:better_algo_tab} contains the results and associated ablation studies for \textbf{Experiment 1}. The details of the ablation studies are explained below:

\begin{enumerate}
    \item \textbf{Random} has a 3D point randomly generated within the range of [max, min] of the training set's labels
    
    \item \textbf{EgoPAT3D\_VF} and \textbf{EgoPAT3D} are two baseline algorithms from the original EgoPAT3D paper \cite{li2022egocentric}. \textbf{EgoPAT3D\_VF} uses only point cloud as input, while \textbf{EgoPAT3D} additionally takes in the IMU data. Note that, as mentioned in Sec. \ref{subsubsec:dataset} and Sec. \ref{subsubsec:implementation}, our baseline implementations actually perform better than those in the original paper.
    
    \item \textbf{ConvNeXt\_Only} and \textbf{ResNet50\_Only} simply replace the PointConv \cite{PointConv} in the \textbf{EgoPAT3D\_VF} with a pre-trained ConvNeXt\_Tiny in \cite{ConvNext} and pre-trained ResNet50 in \cite{resnet}, respectively. It does not have any of the modifications mentioned in Sec. \ref{sec:method}
    \item \textbf{Post+Hand} does not have the new losses discussed in Sec. \ref{subsec:loss} but has post-processing and hand features.
    
    \item \textbf{Post+$L^{Hand}$+$L^{Time}$} does not have a hand feature extractor, so there is no MLP to fuse the hand features with the vision feature. However, it has the post-processing and the new losses.

    \item \textbf{Hand+$L^{Hand}$+$L^{Time}$} does not have the post-processing mentioned in Sec. \ref{subsubsec:post_processing} but has hand features and the two losses.

    \item \textbf{Post+Hand+$L^{Hand}$} and \textbf{Post+Hand+$L^{Time}$} either Hand Position Loss or Time Loss mentioned in Sec. \ref{subsec:loss} while keeping other components the same as the \textit{\acronym{}} algorithm

    \item \textbf{ResNet101\_Replace} replaces the ConvNeXt\_Tiny in \textit{\acronym{}} with a ResNet101.
    \item \textbf{EgoPAT3Dv2\_Transformer} relaces the LSTM in \textit{\acronym} with a transformer. In our low-data regime, the transformer does not perform as well, which has been observed in similar works \cite{Melis2020Mogrifier, Izsak_efficientDL}.
    \item \textbf{EgoPAT3Dv2\_Full} adds point cloud data and IMU data. They are processed and fused as in \textbf{EgoPAT3D}. This experiment shows that simply providing more information by increasing data modality does not translate to better accuracy.
\end{enumerate}

\textbf{Discussion on RGB Backbones}. As we can see with \textbf{ConvNeXt\_Only} and \textbf{ResNet50\_Only}, simply replacing the PointConv with a CNN network feature extractor can substantially improve the performance, especially late prediction. We hypothesize that 2D RGB input is adequate for simple monocular depth estimation due to the depth cues that come with it and because our manipulation task has limited depth variety. At the same time, feature extractors pre-trained on the massive ImageNet \cite{ImageNet} can be more capable than three PointConv layers trained from scratch. Along with \textbf{ResNet101\_Replace}, we show that ConvNeXt\_Tiny (28,589,128 \#params) performs significantly better than ResNet50 (25,557,032 \#params) and larger ResNet101 (44,549,160 \#params) in our algorithm. The small number of parameters guarantees strong inference speed for real-time tasks. More discussion on performance can be found in Sec.\ref{sec:realworld} \label{dis:backbone}

\textbf{Discussion on Hand Feature}.
One issue not discussed in the original EgoPAT3D baseline \cite{li2022egocentric} is that the visual feature extraction is only at the global hierarchy, as PointConv produces a global feature for each point cloud input. Nothing is explicitly done about the hand manipulating the object, even though prior knowledge tells us that the action target will be highly correlated to the hand's location and movement through time. In our algorithm, hand features are fused into the pipeline, and experiment \textbf{Post+Loss} shows that early-stage prediction performance suffers substantially without the hand features.

\textbf{Discussion on Post-Processing}. \label{dis:post-processing}
One salient issue we observe in the baseline of EgoPAT3D \cite{li2022egocentric} is that the network's accuracy stagnates and degrades when the hand moves towards the action target during the later stages. We experimented using PointConv as a feature extractor to perform hand detection on the very last frame and achieved an error of less than 0.2 cm. This experiment shows that the LSTM network does not understand when the manipulation task is approaching the end. To tackle this issue, we adopt a post-processing procedure that considers hand movement. We observe that hand movement tends to stabilize toward the end of a manipulation task so one can place an item stably. Therefore, in our post-processing, we put more weight on the hand detector when the stabilization of hand locations is detected, and we trust the LSTM's "raw" output more when it is in the early stages and the hand is still moving fast. We can see in \textbf{Hand+$L^{Hand}$+$L^{Time}$} that late-stage performance suffers significantly when there is no post-processing.

\textbf{Discussion on New Losses}. \label{dis:losses}
The \textbf{Hand Location Loss} tries to tackle the same problem as the Hand Feature input. Visual features should not be simple global features but focus on the hand. The \textbf{Time Loss} is trying to solve a similar problem as post-processing. We want the LSTM network to pay more attention to the progression of a manipulation task without introducing a hardcoded positional encoding that will be at odds with our online prediction setting. By observing \textbf{Post+Hand}, \textbf{Post+Hand+$L^{Hand}$} and \textbf{Post+Hand+$L^{Time}$}, we see that the combination of the two losses can create a substantial performance boost compared to using only one.

\subsubsection{\textbf{Experiment 2}}
Table \ref{tab:seen_tab} contains the results and associated ablation studies for \textbf{Experiment 2}. The details of the ablation studies are explained below.
\begin{table}[h]
	\scriptsize
	\centering
	\caption{\textbf{Prediction error (cm) of \textit{\acronym{}}.} The models are trained separately on the original EgoPAT3D and the enhanced EgoPAT3Dv2 dataset. They are then tested on the seen and unseen test sets of the original EgoPAT3D and enhanced EgoPAT3Dv2 dataset. \textcolor{BrickRed}{Red}, \textcolor{green!60!black}{green}, and \textcolor{blue}{blue} fonts denote the top three performance.}
 	\vspace{-6pt}
 	\renewcommand\tabcolsep{0.5pt}
	\resizebox{0.46\textwidth}{!}{
		\begin{tabular}{l|c|ccccc|c cccc}

\hline
			\multirow{2}{*}{\textbf{Description}}& \multirow{2}{*}{\textbf{Ovr.}} &\multicolumn{5}{c|}{\textbf{Early prediction}} &\multicolumn{5}{c}{\textbf{Late prediction}} \\
			 &   &10\% & 20\%  & 30\% & 40\%& 50\% &60\% & 70\%  & 80\% & 90\%  & 100\% \\
    \hline
    EgoPAT3D\ & 18.8 & 22.8 & 20.4 & 19.3 & 18.6 & 18.0 & 17.6 & 17.3 & 17.00 & 16.9 & 17.0 \\
    
    \hline
			
    T1\_D1\_Seen & \textcolor{BrickRed}{15.0} & \textcolor{green!60!black}{20.4} & \textcolor{BrickRed}{19.0} & \textcolor{BrickRed}{18.3} & \textcolor{BrickRed}{17.4} & \textcolor{green!60!black}{15.7} & \textcolor{green!60!black}{13.7} & \textcolor{green!60!black}{11.6} & \textcolor{green!60!black}{9.7} & \textcolor{green!60!black}{7.5} & \textcolor{BrickRed}{6.6} \\
    T1\_D1\_Unseen & 16.3 & \textcolor{blue}{21.2} & \textcolor{blue}{19.6} & \textcolor{green!60!black}{18.6} & \textcolor{green!60!black}{17.6} & 16.3 & 15.1 & 13.7 & 12.0 & 10.9 & 10.4 \\
    T1\_D2\_Seen & 21.6 & 28.0 & 26.6 & 25.4 & 23.8 & 21.5 & 19.2 & 17.0 & 15.5 & 14.3 & 14.00 \\
    T1\_D2\_Unseen & 21.5 & 24.0 & 22.5 & 21.5 & 20.3 & 18.9 & 17.6 & 15.8 & 14.2 & 13.2 & 12.8 \\
    \hline
    T2\_D1\_Seen & \textcolor{green!60!black}{15.2} & \textcolor{BrickRed}{20.0} & \textcolor{green!60!black}{19.2} & \textcolor{blue}{18.8} & 18.0 & \textcolor{blue}{16.2} & \textcolor{blue}{14.1} & \textcolor{blue}{11.8} & \textcolor{blue}{9.9} & \textcolor{blue}{7.8} & \textcolor{green!60!black}{6.9} \\
    T2\_D1\_Unseen & 16.6 & 21.6 & 20.6 & 19.8 & 18.7 & 17.1 & 15.4 & 13.2 & 11.0 & 10.00 & 9.8 \\
    
    T2\_D2\_Seen & \textcolor{green!60!black}{15.2} & 22.0 & 20.7 & 19.5 & \textcolor{blue}{17.9} & \textcolor{BrickRed}{15.3} & \textcolor{BrickRed}{12.7} & \textcolor{BrickRed}{10.2} & \textcolor{BrickRed}{8.5} & \textcolor{BrickRed}{7.4} & \textcolor{blue}{7.1} \\

    T2\_D2\_Unseen & 17.6 & 22.7 & 21.8 & 20.9 & 19.7 & 18.00 & 16.0 & 13.8 & 12.00 & 11.2 & 11.0 \\
    \hline
	\end{tabular}}
	\label{tab:seen_tab}%
 	\vspace{-20pt}
\end{table}%
\newline
% \textcolor{BrickRed}{}    \textcolor{green!60!black}{}    \textcolor{blue}{}
\begin{itemize}
    \vspace{-0.2cm}
    \item \textbf{EgoPAT3D} is the original baseline of EgoPAT3D evaluated on the unseen part of the EgoPAT3D test set.
    \item \textbf{T1} means the model is trained with the original training set of EgoPAT3D.
    \item \textbf{T2} means the model is trained on our enhanced dataset. The enhanced dataset contains the original EgoPAT3D training set and our additions.
    \item \textbf{D1} means the model is evaluated on the test set of the original EgoPAT3D.
    \item \textbf{D2} means the model is evaluated on the test set of our enhanced dataset, which contains the original EgoPAT3D's test set and our additions.
    \item \textbf{Seen} means that the model is evaluated on the seen part of the test set, as described in Sec. \ref{subsubsec:dataset}
    \item \textbf{Unseen} means that the model is evaluated on the unseen part of the test set, as described in Sec. \ref{subsubsec:dataset}
\end{itemize}
As we can see in \textbf{T1\_D1\_Unseen}, \textit{\acronym{}} also performs significantly better on the unseen test set of the original EgoPAT3D dataset compared to the original baseline.

Through all the models with \textbf{T2} labels, we can see when trained with additional data, our best model maintains a competitive performance when evaluated on the original EgoPAT3D dataset, on both seen and unseen scenes. At the same time, our model with additional training data can achieve better generalization when facing a more extensive and diverse test set.

\vspace{-0.2cm}
\section{Dataset Enhancement}\label{sec:dataset}
\vspace{-0.2cm}
\subsection{\textbf{Size}}
\begin{figure}
    \centering
    \captionsetup{type=figure}
    \includegraphics[width=\linewidth]{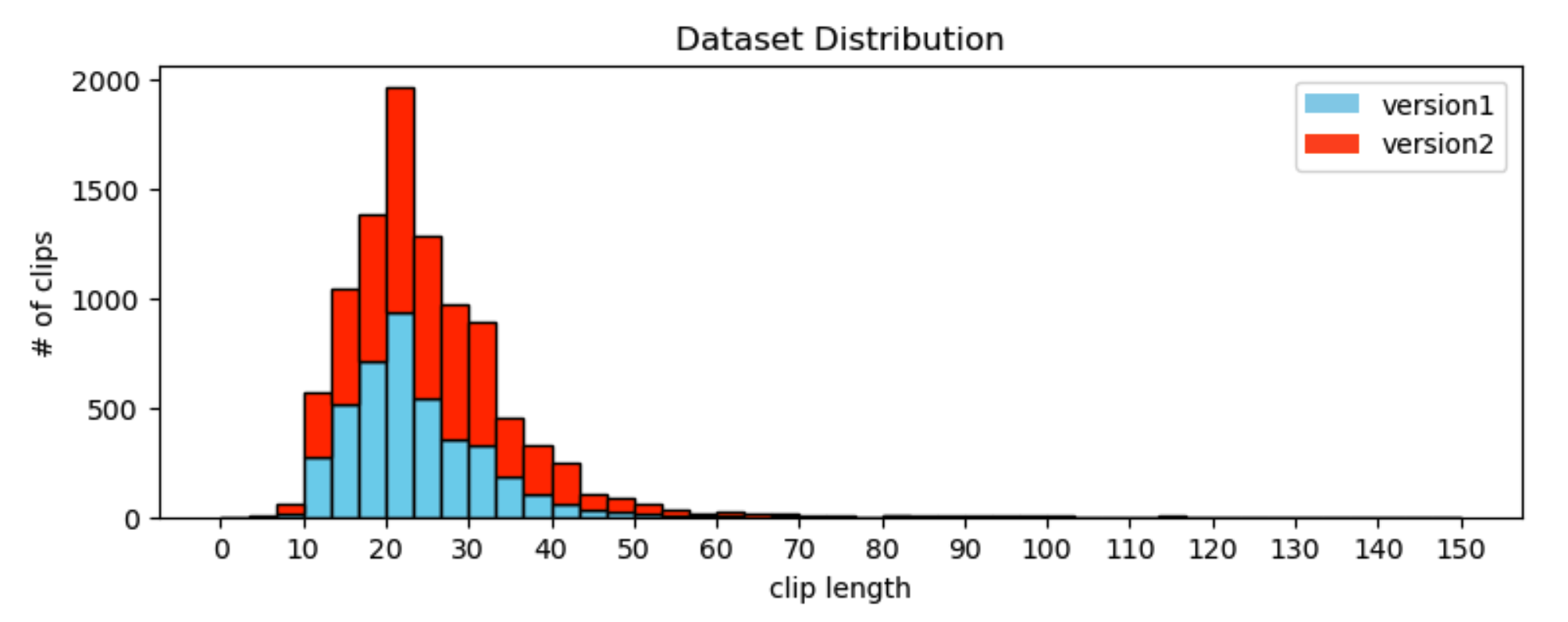}
    \captionof{figure}{Distribution of the clip length in EgoPAT3D and EgoPAT3Dv2 dataset}
    \label{fig:DatasetComparison}
    \vspace{-0.7cm}
\end{figure}

\vspace{-0.2cm}
We doubled the size of the EgoPAT3D dataset. The total number of available clips increases from 4129 to 9579. The seen test set of the new data has clip lengths ranging from 10 to 115 frames, with a mean of 24. The unseen test set runs from 8 frames to 42 frames, with an average of 20. The distribution of clip length can be seen in Fig. \ref{fig:DatasetComparison}
\vspace{-0.25cm}
\subsection{\textbf{Diversity}}
\vspace{-0.25cm}
We have nine additional individuals as human subjects, compared to 2 in the first version. The volunteers are from different countries with different skin complexions.

We also added 12 new scenes that are substantially different from those in the original EgoPAT3D dataset, with a diverse array of objects for humans to manipulate.
\vspace{-0.2cm}
\subsection{\textbf{Data Collection and Annotation}}
\vspace{-0.15cm}
The data collection process follows the same procedure in the original EgoPAT3D paper \cite{li2022egocentric} with a Microsoft Azure Kinect camera mounted on a helmet.

Instead of the semi-automatic data annotation process described in \cite{li2022egocentric}, we opted for a more reliable manual annotation method. While we still use MediaPipe \cite{MediaPipe} for hand detection, we have humans to verify the results and manually annotate if MediaPipe fails to perform the detection.

% \twocolumn[{%
% \renewcommand\twocolumn[1][]{#1}%
% \maketitle
% \vspace{-10mm}
% \begin{center}
%     \centering
%     \captionsetup{type=figure}
%     \includegraphics[width=\textwidth]{ICRA2024/figs/dataset_distribution_comparison.png}
%     \captionof{figure}{\textbf{Real-World Demo of Our Algorithm}. }
%     \label{fig:fig1}
% \end{center}%
% }]

\section{Real-World Demonstration} \label{sec:realworld}
We deployed our algorithm on a UR10e robot for simple demonstrations that reflect a real-world scene where a human and a robot share a common workspace. In setting \#1, if the predicted 3D action target enters a ball of radius of $r$ around the robot's end-effector, the robot reactively moves away to avoid collision with the human. In setting \#2, the robot tries to reach the target position with the shortest Cartesian path.

A hand-eye calibration using MoveIt \cite{MoveIt} is performed as an eye-to-hand setup. The motion planning is done through MoveIt and OMPL \cite{sucan2012the-open-motion-planning-library}.

In a PC with i7-13700F and RTX 4070 Ti, the algorithm can achieve a stable 17 FPS thanks to the small parameters count of ConvNeXt\_Tiny.

The demo videos can be found on our project webpage. A frame-by-frame excerpt can be found in Fig. \ref{fig:fig1}

\vspace{-0.1cm}
\section{Conclusion and Limitation}
\vspace{-0.1cm}
In this work, we greatly improve the 3D human action target prediction performance on the EgoPAT3D \cite{li2022egocentric} dataset. Our algorithm incorporates prior knowledge about manipulation into the learning process and also utilizes state-of-the-art vision pre-training to reduce the required number of sensing modalities. We also enhance the original EgoPAT3D dataset and improve its diversity. Finally, we deploy the algorithm on a real robot to demonstrate the potential of 3D human action prediction for human-robot interaction (HRI).

However, we realize the \textit{limitations in our real-world demonstration, algorithm, and dataset.}
The complexity and variety of our real-world demonstration is limited. Complex tasks with both hands visible should be attempted to investigate the robustness of the algorithm. The variety of robotic platforms in our HRI demonstration can also be expanded (e.g., wearable robots, robotic prostheses) since we want a generic egocentric vision algorithm. 

The dataset and algorithm can also be further improved. Our current dataset and algorithm assume that there will only be one single hand throughout most of the videos, limiting its real-world use cases. The improvement of early-stage performance is also limited compared to late stages.

%%%%%%%%%%%%%%%%%%%%%%%%%%%%%%%%%%%%%%%%%%%%%%%%%%%%%%%%%%%%%%%%%%%%%%%%%%%%%%%%

\clearpage
{\small
\bibliographystyle{IEEEtranN}
\balance
\bibliography{reference}
}

\end{document}